\documentclass[runningheads]{llncs}
\usepackage{graphicx}

\usepackage{tikz}
\usepackage{comment}
\usepackage{amsmath,amssymb} 
\usepackage{color}
\usepackage{ulem}
\usepackage{multirow}
\usepackage{marvosym}

\usepackage[accsupp]{axessibility}  

\begin{document}
\pagestyle{headings}
\mainmatter
\def\ECCVSubNumber{6862}  

\title{D2-TPred: Discontinuous Dependency for Trajectory Prediction under Traffic Lights} 

\titlerunning{D2-TPred: Discontinuous Dependency Trajectory Prediction}
%
\author{Yuzhen Zhang\inst{1} \and
Wentong Wang\inst{2} \and
Weizhi Guo\inst{1} \and
Pei Lv\inst{1}{\textsuperscript{\Letter}} \and
Mingliang Xu\inst{1} \and \\ 
Wei Chen\inst{3} \and
Dinesh Manocha\inst{4}
}

\authorrunning{Y. Zhang et al.}
%
\institute{School of Computer and Artificial Intelligence, Zhengzhou University, Zhengzhou, China. \email{zyzzhang@gs.zzu.edu.cn,\\ \{ielvpei, iexumingliang\}@zzu.edu.cn, ieguoweozhi@163.com} \and
Henan Institute of Advanced Technology, Zhengzhou University, Zhengzhou, China.
\email{wangwentong@gs.zzu.edu.cn}\\
\and State Key Lab of CAD\&CG, Zhejiang University, Zhejiang, China.
\email{chenwei@cad.zju.edu.cn}\and Department of Computer Science, University of Maryland, College Park, MD, USA.\\
\email{dmanocha@umd.edu}
}

\maketitle

\begin{abstract}
A profound understanding of inter-agent relationships and motion behaviors is important to achieve high-quality planning when navigating in complex scenarios, especially at urban traffic intersections. We present a trajectory prediction approach with respect to traffic lights, D2-TPred, which uses a spatial dynamic interaction graph (SDG) and a behavior dependency graph (BDG) to handle the problem of discontinuous dependency in the spatial-temporal space. Specifically, the SDG is used to capture spatial interactions by reconstructing sub-graphs for different agents with dynamic and changeable characteristics during each frame. The BDG is used to infer motion tendency by modeling the implicit dependency of the current state on priors behaviors, especially the discontinuous motions corresponding to acceleration, deceleration, or turning direction. Moreover, we present a new dataset for vehicle trajectory prediction under traffic lights called VTP-TL. Our experimental results show that our model achieves more than  {20.45\% and 20.78\% }improvement in terms of ADE and FDE, respectively, on VTP-TL as compared to other trajectory prediction algorithms. The dataset and code are available at: \url{https://github.com/VTP-TL/D2-TPred}.

\end{abstract}

\section{Introduction}

\begin{figure}[ht]
\centering
\includegraphics[height=5.5cm]{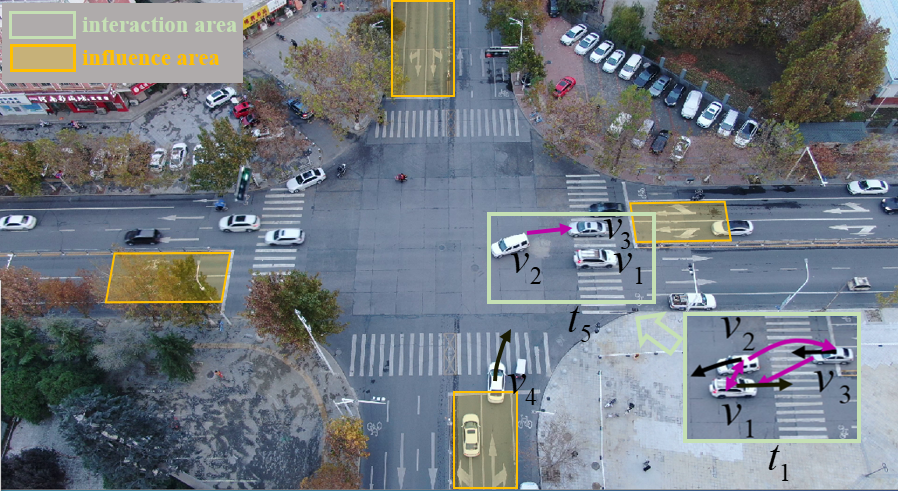}
\caption{Illustration of discontinuous dependency among vehicles at crossroad intersection near traffic lights. We highlight the local trajectories of four vehicles, $v_1 \ldots v_4$ using black directed curves. The orange boxes represent the influence area of the corresponding traffic lights, which are fixed regions and restrict the motion behavior of vehicles while they are passing. We show the position of three vehicles, $v_1, v_2, v_3$ at time $t_1$ and $t_5$, along with the corresponding green boxes which show the dynamic interaction area determined by the moving vehicles. The purple directed edges within each green box represent the interactions among vehicles. In this case, $v_3$ interacts with $v_1$ and $v_2$ at time $t_1$. However, $v_3$ is not affected by $v_1$ at time $t_5$, even though it is located in the same region. This indicates the discontinuity in the interaction between $v_1$ and $v_3$  during this time period. The vehicle $v_4$ located in the influence areas is not constrained by the red light because it is right-turning.}
\label{fig1}
\end{figure}

The interaction relationships and behavioral intentions of vehicles or agents are frequently used for autonomous driving~\cite{2015Intentionaware,2018PORCA,2016MotionPlanning}. A key problem is to predict the future trajectory of each vehicle or road agent, which is used to perform safe navigation or traffic forecasting~\cite{2018Socialgan,2021AG-GAN,2019TraPHic,2021Tra2Tra}. 
Existing trajectory prediction methods are mainly designed to extract the spatial-temporal information from spatial interactions and behavior modeling. In terms of spatial interaction, most previous works determine the interaction among objects according to the predefined interaction areas, such as the entire scene~\cite{2018Socialgan,2021AG-GAN,2020SoPhie,2018SocialAttention,2021Tra2Tra}, localized regions~\cite{2016Social,2018ConvolutionalSocial,2019TraPHic}, and the area corresponding to visual attention~\cite{2021ForecastingPeople}. However, these methods do not fully consider the varying interactions and dependency between neighbors that occur due to different behaviors, such as changing lanes or turning directions that can lead to new pairwise interactions. In terms of behavior dependency, these prediction algorithms obtain the relevant information of the current state from previous states based on LSTM-based methods~\cite{2019STGAT,2021Tra2Tra} or graph-based approaches~\cite{2020SocialSTGCNN,2021SGCN}. 

In this paper, we address the problem of trajectory prediction in areas close to traffic lights or intersections. Due to the constraints of traffic signs and traffic lights with red, green, and yellow states labeled by discrete indexes, the vehicles usually do not exhibit the first-order continuity in their movement behaviors with stopping, going straight, turning right, and turning left. Instead, their trajectory is governed by the discontinuous effects from the environment or other agents. For example, in the green boxes of Figure~\ref{fig1}, the interactions among vehicle $v_{1}$, $v_{2}$ and $v_{3}$ change from time $t_{1}$ to $t_{5}$. Even though these vehicles are within the same interaction regions determined by distance which shown using green boxes, the spatial and behavior interaction between the vehicles changes considerably and we need to model such changes. For vehicle $v_{4}$, the most important influence on its current state is the change in its behavior due to the right-turn, rather than the movement state in adjacent timestamp. We refer to these phenomena as \textit{discontinuous dependency  (D2)}, which makes accurate spatial-temporal feature extraction extremely challenging. Current trajectory prediction methods do not fully account for this property that the trajectories of traffic agents are usually not first-order continuous due to the frequent starting and stopping motions.

\noindent{\bf Main Results:} In order to model the discontinuous dependency between traffic agents, we present a new trajectory prediction approach (D2-TPred). In our formulation, we construct a spatial dynamic interaction graph (SDG) for different traffic agents in one frame. Each traffic agent is regarded as a graph node and we compute appropriate edges to model its interactions with other changing neighboring agents determined by visual scope, distance, and lane index as well as discontinuous dependencies in terms of their relative positions. Moreover, a behavior dependency graph (BDG) is computed for each agent to model the discontinuities with respect to their behaviors at previous time instances, rather than only adjacent timestamp. Specifically, to avoid the key behavioral features such as acceleration, deceleration, or turning direction, may be filtered by forget gates or the error will be accumulated in sequential prediction by RNN network, the way of dependency information passing between adjacent frames is replaced by a GAT (graph attention network)~\cite{2017GraphAttention}, and the behavior dependency is modeled along the edges in the BDG. The SDG and BDG are used as part of a graph-based network for trajectory prediction.

We also present a new dataset for vehicle trajectory prediction, VTP-TL. Our dataset consists of traffic pattern at urban intersections with different traffic rules, such as crossroads, T-junctions intersections, and roundabouts, containing 2D coordinates of vehicle trajectory and more than 1000 annotated vehicles at each traffic intersection. The novel components of our work include:

\begin{itemize}
    \item [1.]
    We propose a novel trajectory prediction approach, D2-TPred, that accounts for various discontinuities in the vehicle trajectories and  pairwise interactions near traffic lights and intersections.  
    \item [2.] We present two types of data structure to improve the performance of graph-based networks to model dynamic interactions and vehicle behaviors. SDG is used to model spatial interactions by reconstructing appropriate sub-graphs for dynamic agents with constantly changing neighbors in each frame. BDG is used to model the dynamically changing behaviors dependency of current state on previous behaviors. The usage of SDG and BDG improves the prediction accuracy by 22.45\% and 29.39\% in ADE and FDE.
    \item [3.]
    We present a new dataset VTP-TL that corresponds to traffic video data near traffic lights and interactions. This includes $150$ minutes of video clips at $30$fps corresponding to challenging urban scenarios. They are captured using drones at $70-120$ meters above the traffic intersections.
\end{itemize}

\section{Related Work}
A brief overview of prior work on graph neural networks, interaction models, and motion pattern dependency is given.

\textbf{Graph Neural Networks:} Graph Neural Network (GNN)~\cite{2018Graph} can model social or other interactions between agents. Prior trajectory prediction methods based on GNN can be divided into two categories. The first is based on undirected graphs, which utilize the graph structure to explicitly construct interactions and assign the same weight for each pair of nodes, e.g., STUGCN~\cite{2021STUGCN}, Social-STGCNN~\cite{2020SocialSTGCNN}. The second is based on Graph attention networks (GAT)~\cite{2017GraphAttention}, which introduces an attention mechanism into the undirected graph to calculate asymmetric influence weights for interactive agents. The GAT-based approaches, such as Social-BiGAT~\cite{2019SocialBiGAT}, STGAT~\cite{2019STGAT}, EvolveGraph~\cite{2020EvolveGraph}, and SGCN~\cite{2021SGCN}, can flexibly model asymmetric interactions to compute spatial-temporal features and improve the prediction accuracy. Meanwhile, EvolveGraph~\cite{2020EvolveGraph} and SGCN~\cite{2021SGCN} introduce
graph structure inference to generate dynamic and sparse interaction. Different from these methods, we directly construct one directed graph according to interactive objects determined by the visual scope, distance, and traffic rules, and use GATs to represent the asymmetric interactions among agents.

\textbf{Social Interaction Models:} Social interactions and related information are used by traffic agents to make reasonable decisions to avoid potential collisions. Social force-based methods~\cite{1995Social,2009Abnormal,2018AutoRVO} use different types of force to model acceleration and deceleration forces. Social pooling based approaches~\cite{2016Social,2018ConvolutionalSocial,2018Socialgan,2021AG-GAN} try to integrate motion information of neighbors within a radius. GNN-based techniques~\cite{2018SocialAttention,2019SocialBiGAT,2020EvolveGraph,2020SocialSTGCNN,2020Trajectron++,2019STGAT,2021AComprehensive} use graph structures to directly model the interactions among near and far agents. These methods assume that the underlying agents interact with all other agents in a predefined or nearby regions. They do not account for those neighbors need to be pruned, especially moving along the opposite lanes.

\textbf{Motion Models:} Motion models are used to infer the motion information as part of trajectory prediction. Early studies have focused on forecasting future trajectories based on a linear model, constant velocity model, or constant acceleration model~\cite{2016Trajectoryof}. However, these simple models can not handle complex traffic patterns. Furthermore, LSTM-based approaches~\cite{2016Social,2018Socialgan,2019STGAT,2021Pedestrian} and graph structure-based approaches~\cite{2019TGCN,2021STUGCN,2020SocialSTGCNN,2021SGCN} are proposed to model the motion trajectories. Other techniques take into account driver behavior patterns~\cite{CMetric2020,B-GAP2020}. Giuliari et al.~\cite{2020TransformerNetworks} perform precise trajectory prediction using transformer networks. Here, the states of an agent in temporal sequence are regarded as nodes to construct directed graph, further achieve the direct influence between discontinuous timestamps, rather than only adjacent one.

\section{D2-TPred}
\begin{figure}[!t]
\centering
\includegraphics[width=0.75\linewidth]{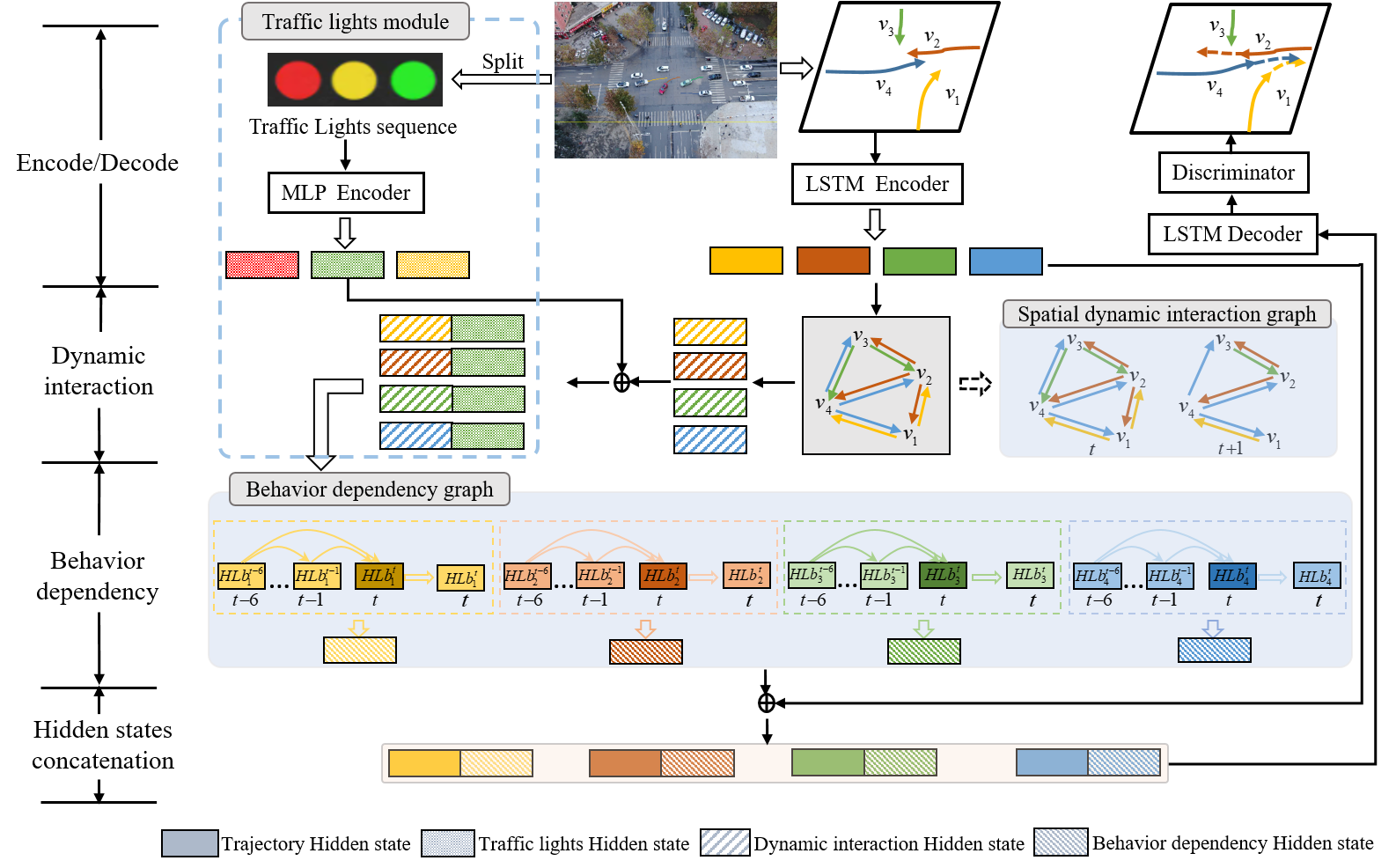}
\caption{Architecture of our proposed D2-TPred model. The spatial dynamic interaction graph is used to represent the dynamic interactions and  we reconstruct the sub-graphs. The behavior dependency graph learns the movement features by estimating the effect of agent behavior. The discriminator is used to refine the predicted trajectories. The traffic light module is used to predict the trajectories at urban intersections.}
\label{fig2}
\end{figure}

In this section, we present our novel learning-based trajectory prediction algorithm, which involves the influence of traffic lights on motion behaviors, and the architecture is shown in Figure~\ref{fig2}.

\subsection{Problem Formulation}
Given spatial coordinate and traffic light state of $N$ agents in each scenario, we aim to predict the most likely trajectories of these agents in the future. At any time $t$, the state of the $i$th agent $Sq_{i}$ at time $t$ can be denoted as $Sq_{i}^{t} = (Fid,Aid,x_{i}^{t},y_{i}^{t},Lid,pa_{i}^{t},f_{i}^{t},mb_{i}^{t},lid_{i}^{t},ls_{i}^{t},lt_{i}^{t})$, where $p_{i}^{t}=(x_{i}^{t},y_{i}^{t})$ represents the position coordinate and the other symbols represent the corresponding traffic light information described with more detail in Section 3.3. According to the inputs of all agents in the interval $[1: {t}_{obs}]$, our method can predict their position at next moment ${t}_{pred}\in[{t}_{obs}+1: T]$. Different from the ground truth trajectory ${Lq}_{i}^{{{t}_{pred}}}$, $\hat{Lq}_{i}^{{{t}_{pred}}}$ notates the predicted trajectory. 

\subsection{Spatio-Temporal Dependency}

\begin{figure}[ht]
	\begin{center}
		\includegraphics[width=0.75\linewidth]{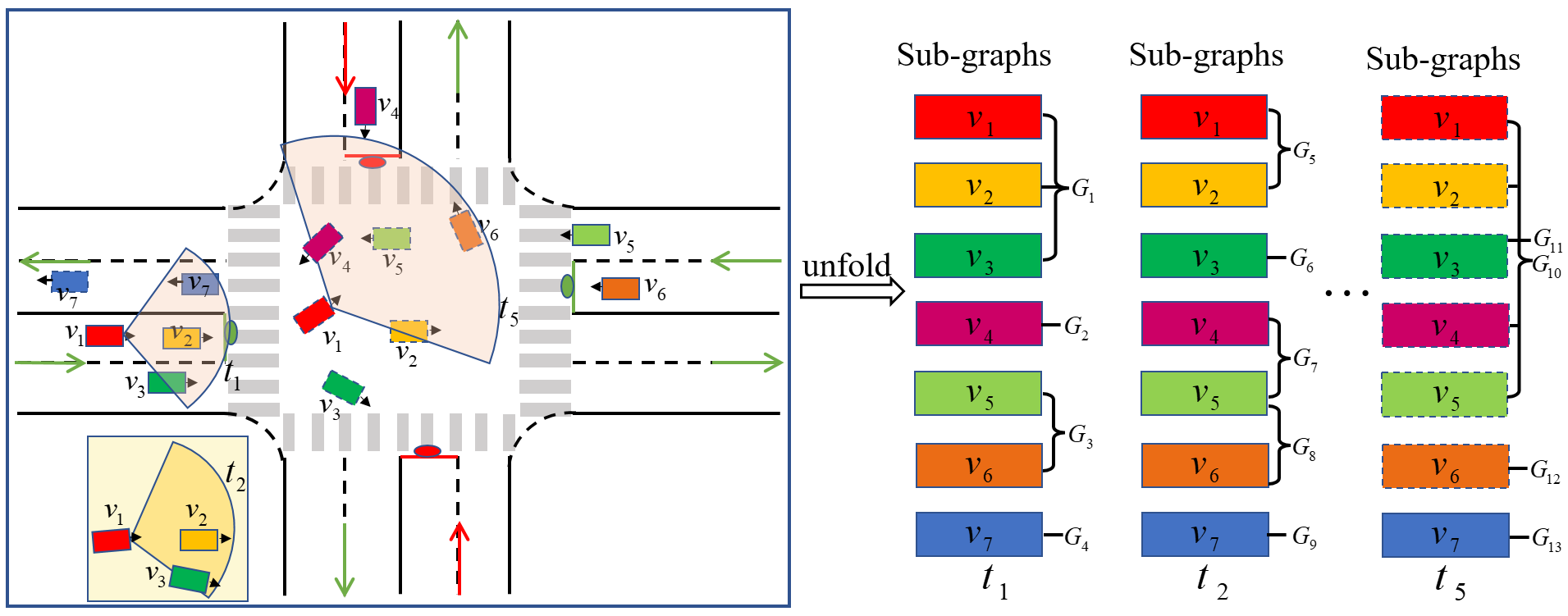}
	\end{center}
	\caption{The spatial dynamic interaction graph (SDG). The left part of the figure shows the scene from time $t_{1}$ to $t_{5}$, and the right part represents the reconstructed interaction sub-graphs at different time instances.}
	\label{fig3}
\end{figure}

\textbf{Spatial Dynamic Interaction Graph.} Unlike prior methods~\cite{2019STGAT,2020SocialSTGCNN}, we reconstruct the sub-graphs to model all the interactions in each frame. We illustrate our approach to model discontinuous dependency by highlighting one scenario with $7$ vehicles and appropriate trajectories in Figure~\ref{fig3}. Similar to~\cite{2021ForecastingPeople}, the visual area of a subject is treated as frustum, where different visual ranges are set between road and intersection by considering the characteristics of human visual system. At time $t_{1}$, $v_{2}$, $v_{3}$, and $v_{7}$ are located in the visual area of neighborhood of $v_{1}$. However, the motion behavior of $v_{1}$ is not affected by $v_{7}$, which moves in the opposite lane. Hence, we construct sub-graph $G_{1}$ corresponding to the interactions among vehicles $v_{1}$, $v_{2}$, and $v_{3}$, and sub-graph $G_{3}$ for vehicles $v_{5}$ and $v_{6}$. Moreover, for vehicles $v_{4}$ and $v_{7}$ without nearby neighbors, we compute sub-graphs $G_{2}$ and $G_{4}$, respectively. Based on these sub-graphs, the intermediate states of these vehicles are updated. Since the interactions between the vehicles change dynamically, vehicle $v_{1}$ is not affected by vehicle $v_{3}$ at time $t_{2}$. Even though they are within the same interaction region determined by distance, the influence of vehicle $v_{3}$ on vehicle $v_{1}$ is not the same between adjacent frames. In this manner, we reconstruct the corresponding sub-graphs $G_{5}$, $G_{13}$ to represent these varying interactions between the vehicles.

Considering the asymmetry of interactions among agents, we use a self-attention mechanism into these constructed directed graphs to model the spatial interactions. For agent $i$ at time $t$, we first determine its interactive objects $j$ according to the visual scope $\theta$, distance $d$, and lane index $Lid$, and the corresponding matrix $V$, $D$, and $L$ respectively.
\begin{align}
  \left\{
  \begin{aligned}
    & V[i,j]=1 , \ \ \ if\  \theta(\vec{ij}) \in \theta, \\
    & D[i,j]=1,  \ \ \ if\  \left\|{p}_{j}^{t}-p_{i}^{t}\right\|_2\leq d, \\
    & L[i,j]=1, \ \ \ if\  Lid_{i}^{t}=Lid_{j}^{t},\\
    & R = V\times D\times L
    \end{aligned}
\right.
\end{align}
where $R$ filled with 0 and 1 represents adjacency matrix among agents, and we further construct sub-graph based on it. We then calculate the spatial state ${hs}_{i}^{t}$ by integrating the hidden states $h_{j}^{t}$ from interactive objects.
\begin{align}
    e_{i}^{t}=\Phi(p_{i}^{t}, W_{p}),\ \  h_{i}^{t}&=LSTM(h_{i}^{t-1}, e_{i}^{t}, W_{l}),
\end{align}
\begin{align}
hs_{i}^{t}=&\sum_{j\in \mathcal{N}_{i}^{t}}(a_{ij}^{t}R_{ij}^{t})h_{j}^{t},
\label{eq:4}
\end{align}
where $\Phi(\cdot)$ is an embedding function, and $e_{i}^{t}$ is the state vector of agent $i$ at time $t$. Similar to method~\cite{2014NeuralAttention}, $a_{ij}^{t}$ represents the attention coefficient of agent $j$ to $i$ at timestamp $t$, $W_{p}$ and $W_{l}$ are embedding matrix and LSTM cell weight.

\begin{figure}[!t]
	\begin{center}
		\includegraphics[width=0.75\linewidth]{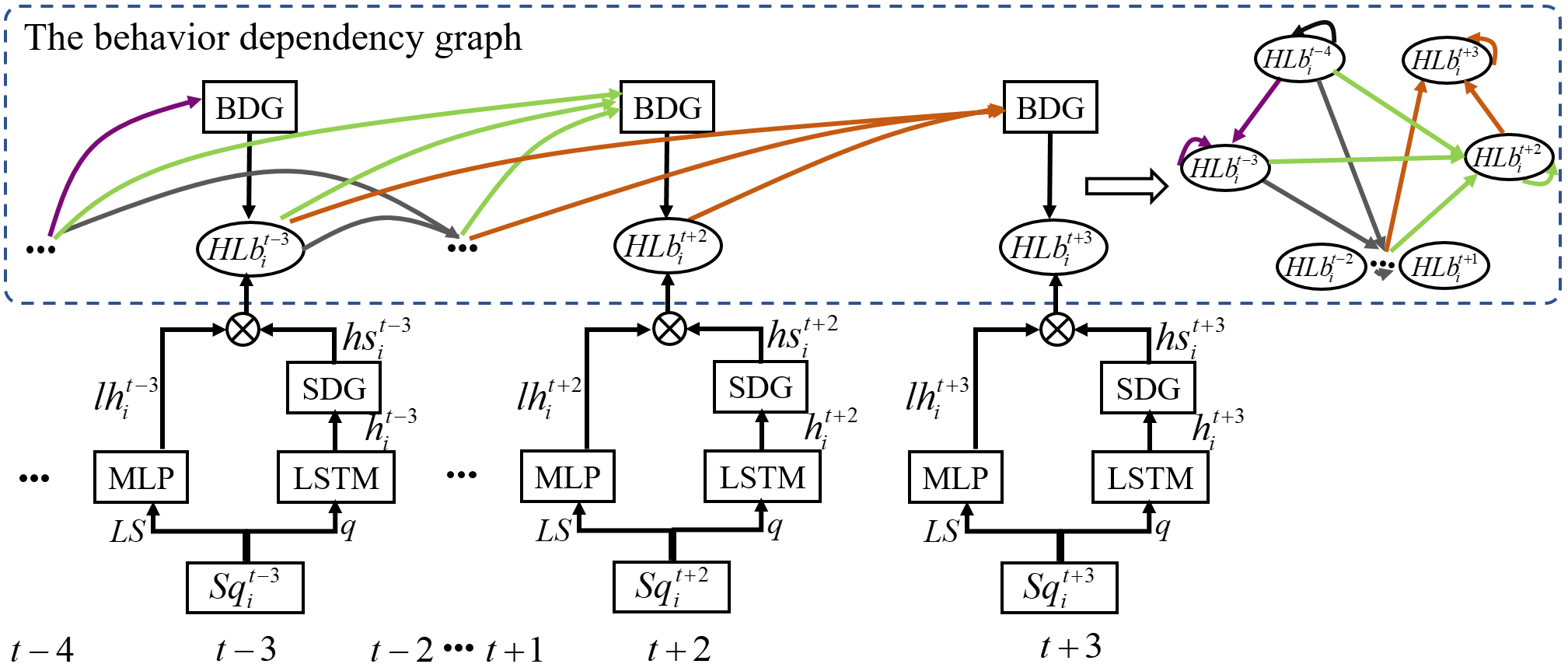}
	\end{center}
	\caption{The behavior dependency graph (BDG). The lower part refers to the encoding process of trajectories and traffic light signals. The upper part describes behavior dependency, where segments with the same color refer to a temporal dependency graph.}
	\label{fig4}
\end{figure}

\textbf{Behavior Dependency Graph.} To avoid the key behavioral features may be filtered by forget gate of RNN network in process of information passing, we model the discontinuous dependency  from previous behaviors to the current state by using GATs, rather than only adjacent timestamps. Specifically, for a given vehicle, its states updated by SDG are regarded as nodes. We model discontinuous dependency in the temporal sequence as edges and construct a directed graph, where the behavior information is transferred along the directed edges. The detailed architecture of the BDG for a given agent is shown in Figure~\ref{fig4}.

Specifically, for agent $i$, we use directed segments with the same color to constitute an unfolded BDG, and different colors represent the behavior dependency graphs at different time instances. The BDG uses the state $hs_{i}^{t}$ generated by the SDG. Its current state is updated and embed into the behavior dependency graph at the next time instance, where dependency weights among nodes are calculated by using a self-attention mechanism. As shown in the dashed box of Figure~\ref{fig4}, the motion state of agent $i$ at current moment $t+2$ is governed by the previous behaviors at time $t+1$, $t$, $t-1$, $t-2$, $t-3$, and $t-4$, etc., whereas the next instance $t+3$ is governed by $t+2$, $t+1$, $t$, $t-1$, $t-2$, and $t-3$. In this way, the updated hidden state $hb_{i}^{t}$ for agent $i$ at time $t$ is calculated as follows:
\begin{align}
\begin{split}
    a_{i}^{tt^{'}}=softm&ax(\frac{exp(LeakyReLU^{*}(\beta^{T}[W{hs}_{i}^{t} \| W{hb}_{i}^{t^{'}}]))}{\sum_{t^{'}}^{t}exp(LeakyReLU^{*}(\beta^{T}[W{hs}_{i}^{t} \| W{hb}_{i}^{t^{'}}]))}), \\ &hb_{i}^{t}=\sum_{t^{'}\in(t-k)\;and\;t^{'}\geq0}^{t}a_{i}^{tt^{'}}(hb_{i}^{t^{'}},\;hs_{i}^{t}),
  \label{eq:4}
\end{split}
\end{align}
where $k$ represents the experimental estimated time window whose quantitative results have the lowest prediction error by obtaining more effective behavior feature. $\beta^{T}$ is the weight vector of a single-layer feedforward neural network.
$t^{'}$ denotes a specific time instance in the previous frames from $t-k$ to $t$.

\subsection{Trajectory Prediction near Traffic Lights}
In this section, we present two prediction schemes for the vehicle trajectory prediction. The first scheme considers the discontinuous constraints on vehicles' behaviors caused by the alternation of traffic light states, where the traffic lights are regarded as indicator signals with fixed position and alternating states. The second scheme is designed for scenarios without traffic lights, which is described in detail in the supplementary material \url{https://github.com/VTP-TL/D2-TPred}.

Given the observed sequence: $Sq_{i}^{t} = (Fid,Aid,x,y,Lid,pa,f,mb,lid,ls,lt)$, which is divided by the vehicle trajectory $q=(Fid, Aid, x, y, Lid, pa, f, mb)$ and the corresponding traffic light states sequence $LS=(Fid, lid, ls, lt)$. $Fid$, $Aid$, and $Lid$ are the index of frame, vehicle, and lane where the vehicle located, respectively. $lid_{i}^{t}$ is the traffic light index. $pa_{i}^{t}$ describes whether vehicle $v_{i}$ is within the influence area of corresponding traffic light. $f_{i}^{t}$ indicates whether vehicle $v_{i}$ is closest to the parking line in the influence area. $mb_{i}^{t}$ represents the movement behavior of one agent, such as turning-left, turning-right, or going-straight. $ls_{i}^{t}$ and $lt_{i}^{t}$  respectively describe the state and duration of traffic light. We take into account that the vehicle trajectory is continuous and the traffic light states sequence is periodic and discontinuous. Therefore, the two different encoders, LSTM and MLP, are utilized to handle them and compute the corresponding hidden states $h_{i}^{t}$ and $lh_{i}^{t}=MLP(LS_{i}^{t}, W_{M})$, respectively. In SDG, we use GAT to integrate influencing features from nearby interacting agents and then compute the updated state ${hs}_{i}^{t}$ of agent $i$. In terms of behavior dependency, we first concatenate the state ${hs}_{i}^{t}$ (Eq.~\ref{eq:4}) and traffic lights state $lh_{i}^{t}$ as input $\tilde{HL}_{i}^{t}$, and then use these results to construct BDG. Based on the BDG, we can model discontinuous constraints of traffic lights on the movement behaviors of vehicles, as shown in Figure~\ref{fig4}. In this stage, hidden state ${HLb}_{i}^{t}$ is computed as a weighted sum of ${HLb}_{i}^{1:(t-1)}$, where the dependency weights are calculated by a self-attention mechanism. The resulting equations:
\begin{align}
  \tilde{HL}_{i}^{t}=integrate({hs}_{i}^{t},\;lh_{i}^{t},\;W_{l}),\ \  HLb_{i}^{t}=GAT(\tilde{HL}_{i}^{t},\;HLb_{i}^{1:t},\;W_{\tilde{\theta}}),
\end{align}
where $integrate(\cdot)$ is concatenation operation. $W_{l}$, and $W_{\tilde{\theta}}$ are the embedding weights. To augment behavior feature and avoid the feature loss filtered by forget gates in process of sequence, the intermediate state are generated by integrating the sate $HLb_{i}^{t}$ and original state $h_{i}^{t}$. The predicted position is given by:
\begin{equation}
  \hat{Lq}_{i}^{t_{pred}}=\sigma(D_{LSTM}([HLb_{i}^{t}, h_{i}^{t}], W_{d})).
\end{equation}
where $D_{LSTM}$ and $W_{d}$ are the decoder based LSTM and corresponding weight respectively. $\sigma(\cdot)$ represents a linear layer. Our method is also GAN-based model integrating a discriminator $D_{cls}$ into the predicted approach, which utilizes LSTM and MLP to respectively encode the complete trajectory ([$Sq_{i}^{T_{obs}}$, $\hat{Lp}_{i}^{T_{pred}}$)] and traffic light sequence $LS$, and then concatenate them as the input into the \textit{Discriminator} to output a real/fake probability $probL_{i}$ by by a linear network.
\begin{align}
 probL_{i}=D_{cls}(LSTM([Sq_{i}^{t_{obs}},\hat{Lq}_{i}^{t_{pred}}], W_{l}), MLP(LS_{i}, W_{M}))
\end{align}

For each vehicle, we calculate displacement error by variety loss in ~\cite{2018Socialgan}. The model predicts multiple trajectories $K$, and chooses the trajectory with the lowest distance error between them and ground-truth trajectory as the model output.
\begin{align}
 Loss_{variety}=\min_{K}\left\|{Lq}_{i}-\hat{Lq}_{i}^{K}\right\|_2.
\end{align}

Through considering the best trajectory, the loss encourages network to cover the space of outputs that conform to the past trajectory.

\subsection{VTP-TL Dataset}
\begin{table}[!t]
  \scriptsize
	\begin{center}
 	\caption{ VTP-TL vs other state-of-the-art traffic datasets. $Size$ represents the number of annotated frames. \textbf{$E_{V}$} and \textbf{$B_{V}$} represent the egocentric vision and bird's view.}
 	\label{table4}
 	\setlength{\tabcolsep}{0.9mm}{
 		\begin{tabular}{l c c c c c c}
 		      \hline
 			 \cline{1-7} Datasets & Location & View & Night & Road type & Size & Traffic lights \\
 			 
 			 \cline{1-7} CityScapes~\cite{Cityscapes2016} & Europe & $E_{V}$ & $\times$ & urban & 25K & $\times$ \\
 			 
 			 Argoverse~\cite{2019Argoverse} & USA & $E_{V}$ & $\surd$ & urban  & 22K & $\times$ \\
 			  INTERACTION~\cite{2019INTERACTION} & International & $B_{V}$ & $\times$ & urban  & - & $\times$ \\
 			 
 			 ApolloScape~\cite{2019TrafficPredict} & China & $E_{V}$ & $\surd$ & urban + rural  & 144K & $\times$ \\
 			 
 			 TRAF~\cite{2019TraPHic} & India & $E_{V}$ & $\surd$ & urban + rural & 72K & $\times$ \\
 		   
 			 D2-city~\cite{2019D2City} & China & $E_{V}$ & $\times$ & urban & 700K & $\times$ \\
 		   
 		    inD~\cite{2019inDdataset} & Germany & $B_{V}$ & $\times$ & urban  & - & $\times$ \\
 			 
 			  Lyft Level5~\cite{2020Olyft5} & USA & $E_{V}$ & $\times$ & urban  & 46K & $\times$ \\
 			 
 		   nuScenes~\cite{2020nuScenes} & USA/Singapore & $E_{V}$ & $\surd$ & urban  & 40K & $\times$ \\
 		   
 		    Waymo~\cite{2021Large} & USA & $E_{V}$ & $\surd$ & urban  & 200K & $\surd$ \\
 
 		   Waterloo~\cite{2021uwaterloo} & Canada & $B_{V}$ & $\times$ & urban  & - & $\surd$ \\
 		   
 		   IDD~\cite{IDD2019} & India & $E_{V}$ & $\times$ & urban + rural & 10K & $\times$ \\
 			 
 			  METEOR~\cite{Meteor2021} & India & $E_{V}$ & $\surd$ & urban + rural & 2027K & $\times$ \\
 			 \hline
 			 \hline
 			 \textbf{VTP-TL} & China & $B_{V}$ & \textbf{$\surd$} & \textbf{urban} & \textbf{270K} & \textbf{$\surd$} \\
			
 			\hline
 		\end{tabular}}
 	\end{center}
 \end{table}

Although plenty of datasets have been constructed to evaluate the performance of trajectory prediction (Table~\ref{table4}), they rarely contained the important attributes of traffic lights except for Waymo and Waterloo. More details and their differences are described in our supplementary materials. For our new traffic datset, VTP-TL, we use drones to hover at $70$ to $120$ meters above the traffic intersections in an urban setting, as statically as possible, to record vehicle trajectories passing through the area with a bird’s-eye view in the daytime corresponding to non-rush hours, rush hours, and during the evening. The dataset contains more than $150$ minutes video clips,
over $270k$ annotated frames,
and more than 4 million bounding boxes for traffic vehicles at three typical urban scenarios: crossroads, T-junctions, and roundabouts intersections. Specifically, according to these recorded videos including traffic lights when they are visible, we infer the invisible state in the videos since the pattern of traffic lights are fixed. Therefore, we manually mark the related attributes of traffic light signals as discrete index, such as the index, state, duration, and position coordinate, and high definition maps such as lanes, crosswalks, and stop lines. Compared with Waymo, the traffic lights are annotated as independent objects, and modeled as agents with fixed position and changeable states in our prediction framework. The center of bounding box is regarded as the position coordinates of vehicle. Finally, we obtain a new dataset of vehicles trajectory prediction containing over $1288$ vehicles driving straight, $801$ vehicles turning left, and $2584$ vehicles turning right. Our dataset is divided into training, validation, and testing sets at a ratio of 4:1:1, and down-sampled to 3 frames per second for experiments.

\begin{figure}[!t]
	\begin{center}
		\includegraphics[width=0.85\linewidth]{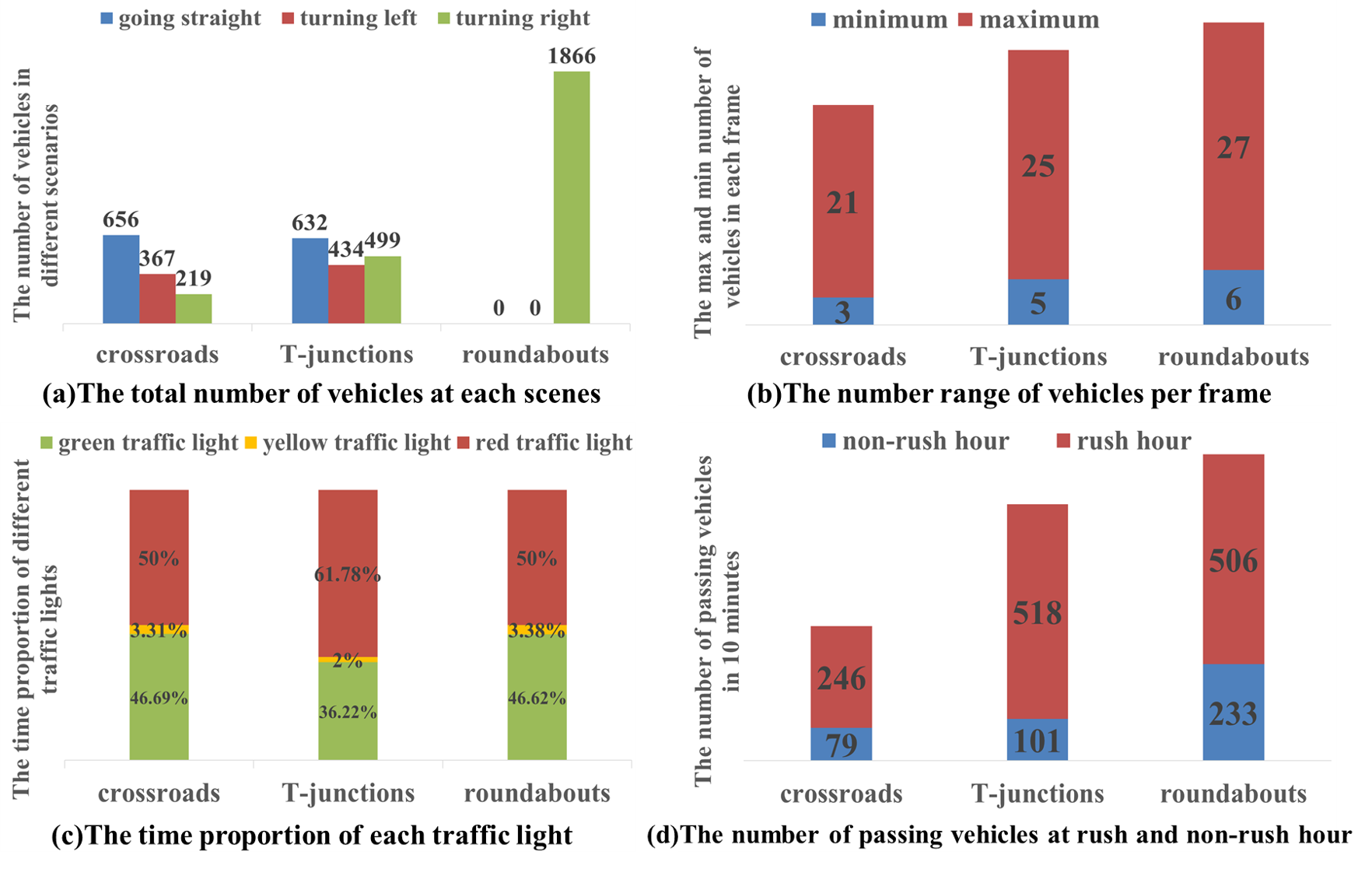}
	\end{center}
	\caption{We highlight the traffic light states and vehicles behaviors in various videos in VTP-TL. The detailed descriptions are given in the appendix.}
	\label{fig7}
\end{figure}

We also perform statistical analysis on the VTP-TL dataset, and the corresponding results are shown in Figure~\ref{fig7}. The number of vehicles with different motion behaviors at each urban intersection are shown in Figure~\ref{fig7}(a). Considering different traffic rules at various urban intersections, we also count the number range of passing vehicles per frame as Figure~\ref{fig7}(b). Meanwhile, to ensure the effective passing of vehicles, different cycle time for traffic lights is set at different intersections (shown in Figure~\ref{fig7}(c)). In Figure~\ref{fig7}(d), we count the number of vehicles in the daytime of the rush and non-rush hours in $10$ minutes. These descriptions can fully represent a large number of vehicles' behaviors are constrained by traffic lights. The user identifiers and exact date of publication have been masked off to protect privacy. The dataset would only be available for research purposes. More details are given in the supplementary materials.

\section{Experimental Evaluation}
In our experiments, the dimension of the embedding layer and the hidden state are set as 16 and 32, respectively. We also set the fixed input dimension as 64 and use the attention layer of 64. During training, the Adam optimizer is applied with a learning rate of 0.01 and batch size of 64.

\textbf{Evaluation Datasets.}
We evaluate proposed model on four traffic datasets,
Apolloscape~\cite{2019TrafficPredict}, SDD~\cite{2016LearningSDD}, INTERACTION~\cite{2019INTERACTION}, and Waymo~\cite{2021Large}.
In addition, we also report the experiments on our new dataset VTP-TL. 

\textbf{Evaluation Metrics.} We use the same evaluating metrics as ~\cite{2019STGAT,2020Trajectron++,2020Collaborative}. \textit{Average displacement error} (ADE) represents the average square error between the predicted trajectory and the ground truth trajectory for all agents at all frames. \textit{Final displacement error} (FDE) represents the mean distance between the predicted path and ground truth trajectory for all agents at the final frame.

\textbf{Comparable methods.} \textit{Social-LSTM}~\cite{2016Social} models spatial interaction by pooling mechanism. \textit{CS-LSTM}~\cite{2018ConvolutionalSocial}, \textit{TraPHic}~\cite{2019TraPHic} and \textit{GRAPH-LSTM}~\cite{2020ForecastingTrajectory} combine CNN with LSTM to perform trajectory prediction. \textit{SGAN}~\cite{2018Socialgan}, SGCN~\cite{2021SGCN}, and \textit{Goal-Gan}~\cite{2021GoalGAN}
use GAN to model spatial interactions and physical attentions. \textit{Social Attention}~\cite{2018SocialAttention}, AI-TP~\cite{2022AITP}, \textit{Trajectron++}~\cite{2020Trajectron++}, \textit{GRIP++}~\cite{2020GRIP++}, \textit{NLNI}~\cite{2021Unlimited}, \textit{EvolveGraph}~\cite{2020EvolveGraph},
and \textit{STGAT}~\cite{2019STGAT} integrate graph structure and attention mechanism to extract spatial-temporal interaction features. \textit{TPNet}~\cite{2020TPNet} and \textit{DESIRE}~\cite{2017DESIRE},
integrate scene contex into prediction framework. \textit{NMMP}~\cite{2020Collaborative} models the directed interaction with the neural motion message passing strategy. SimAug~\cite{liang2020simaug} is trained only on 3D simulation data to predict future trajectories. LB-EBM~\cite{E19} is a probabilistic model with cost function defined in the latent space to account for the movement history and social context for diverse human trajectories. \textit{HEAT}~\cite{2022HEAT}, \textit{TNT}~\cite{2020TNT}, and \textit{MultiPath}~\cite{2019MultiPath} are used on INTERACTION and reported in~\cite{2020TNT}.

\begin{table}[!t]
\scriptsize
	\begin{center}
	\caption{Quantitative results of prediction performance on traffic datasets. The ADE/FDE are calculated for each dataset. The bold fonts correspond to the best results with the lowest error among predicted 20 possible trajectories for each agent, except INTER (INTERACTION) where the lowest error among predicted 6 possible trajectories. For Waymo, we implement those baseline methods according to their open source code, while the other experimental values of comparison methods are all described in open papers.	\textbf{-} denotes methods have not been validated on those datasets.}
	\label{table1}
	\setlength{\tabcolsep}{0.2mm}{
		\begin{tabular}{l|c||l|c|c||l|c}
			\hline
             Method & \textbf{Apolloscape} & Method & \textbf{SDD} & \textbf{INTER} & Method & \textbf{Waymo}
             \\
			 \hline
			 TPNet~\cite{2020TPNet} & 2.23/4.70 & EvolveGraph~\cite{2020EvolveGraph} & 13.9/22.9 &- & SGAN~\cite{2018Socialgan}  & 6.01/11.40  \\
			 \hline CS-LSTM~\cite{2018ConvolutionalSocial} & 2.14/11.70 & Goal-Gan~\cite{2021GoalGAN} & 12.2/22.1 &- & Social-LSTM~\cite{2016Social}  & 4.05/7.59 
			 \\
			 \hline			
			 G-LSTMS\cite{2020ForecastingTrajectory} & 1.12/2.05 & SimAug~\cite{liang2020simaug} & 10.27/19.71 & - & STGAT~\cite{2019STGAT} & 1.68/3.70
			 \\
			 \hline			 
			 SGAN~\cite{2018Socialgan} & 3.98/6.75 & LB-EBM~\cite{E19} & 8.87/\textbf{15.61} & - & SGCN~\cite{2021SGCN} & 1.02/2.26
			 \\
			 \hline			 
			 TraPHic~\cite{2019TraPHic} & 1.28/11.67 & DESIRE~\cite{2017DESIRE} & 19.3/34.1 & 0.32/0.88 &  & 
			 \\
			 \hline			 
			 NLNI~\cite{2021Unlimited}  & 1.09/1.55 & HEAT~\cite{2022HEAT} & - & \textbf{0.19}/0.66& &
			 \\
			 \hline
			AI-TP~\cite{2022AITP} & 1.16/2.13 & TNT~\cite{2020TNT} & - & 0.21/0.67 & &
			 \\
			 \hline
			GRIP++~\cite{2020GRIP++} & 1.25/2.34 & MultiPath~\cite{2019MultiPath} & - & 0.30/0.99 & &
			 \\
			\hline
			\hline
			D2-TPred & \textbf{1.02}/\textbf{1.69} & D2-TPred & \textbf{8.24}/15.89 & 0.29/\textbf{0.62} & D2-TPred & \textbf{0.85}/\textbf{1.89}
			 \\
			
			\hline
		\end{tabular}}
	\end{center}
\end{table}

\subsection{Quantitative Evaluation}
We have performed the detailed quantitative evaluation. On traffic datasets Apolliscape, SDD, INTERACTION, Waymo, and VTP-TL, the quantitative results of prediction performance for D2-TPred and other trajector prediction methods are shown in Table~\ref{table1} and Table~\ref{table3}.

\textbf{Traffic datasets without traffic lights:}
Taking benefits from the SDG and BDG to extract spatio-temporal features, our method achieves competitive performance in the datasets shown in Table~\ref{table1}. Specifically, the performance of our method significantly outperforms comparative methods on Apolloscape. In SDD dataset with a large number of different scenarios, we observe the lowest error on ADE and the third-lowest error on FDE, as well as the lowest error on the FDE on INTER (INTERACTION). Moreover, we also achieve the best performance on Waymo Open Motion dataset by observing 8 frames to predict the next 12 frames. These demonstrate that our model can effectively capture the dynamic changeable interaction features and behavior dependency in complex traffic scenarios. More experimental results on other datasets such as ETH-UCY, Argoverse, nuScenes and inD are described in our supplementary materials.

\begin{table}[!t]
\scriptsize
	\begin{center}
	\caption{Quantitative results on VTP-TL dataset. We compare with the baseline methods and compute the ADE and FDE metrics by using 8 time steps to predict 12 future frames. +TL represents that traffic light states is embedded into the trajectory prediction system. The bold fonts correspond to the best results, which are the lowest error among predicted 20 possible trajectories for each agent.}
	\label{table3}
	\setlength{\tabcolsep}{1.0mm}{
		\begin{tabular}{l|c|c|c|c|c|c|c}
			\hline
			
			\multirow{2}{*}{Metrics} & \multicolumn{7}{|c}{Comparable models (in pixels)} \\
			
			 \cline{2-8} & Social Lstm & Social Attention & SGAN & STGAT & Trajectron++ & NMMP & D2-TPred\\
			 
			 \hline
			 ADE & 54.328 & 43.648 & 37.63 & 28.279 & 39.01 & 35.15 & \textbf{20.685} \\
			 \hline
			 FDE &112.635&	97.614&	75.35 &	61.762&	118.37 & 70.35 & \textbf{47.296}\\
			 
			 \hline
			 \hline
			 \multirow{2}{*}{Metrics}& \multicolumn{7}{|c}{Comparable models+TL (in pixels)} \\

			  \cline{2-8}& Social Lstm & Social Attention & SGAN & STGAT & Trajectron++ & NMMP & D2-TPred\\
			  \hline
			  ADE & 45.04	& 34.460 & 31.56 & 21.245 & 35.456 & 32.33 & \textbf{16.900} \\
			  \hline
			  FDE & 78.52 & 75.825 & 65.67 & 43.620 & 114.365 & 66.35 & \textbf{34.553}\\
			\hline
		\end{tabular}}
	\end{center}
\end{table}

\textbf{VTP-TL dataset with traffic lights:}
In this section, we describe D2-TPred+TL, which introduces traffic light states into D2-TPred approach. In Table~\ref{table3}, we evaluate our model against comparable methods, and all these methods against themselves
with traffic lights. The experimental results show our method outperforms all other methods on the VTP-TL dataset in terms of ADE and FDE. Notably, compared with STGAT with the lowest prediction error, the ADE and FDE of D2-TPred+TL are reduced by 20.45$\%$ and 20.78$\%$. This illustrates we can effectively model constraints of traffic lights on motion behaviors.

\subsection{Ablation Studies}
We present the ablation studies on VTP-TL with traffic light. This not only demonstrates the significance of each component but also highlights the benefits of modeling discontinuity due to traffic lights on vehicle movement behavior. 

\textbf{Evaluation of the SDG and BDG:}
To show the effectiveness of the SDG and BDG, we compare $S_{G}+B_{B}+TL_{M}+D$, $S_{S}+B_{L}+TL_{M}+D$ with $S_{S}+B_{B}+TL_{M}+D$ in Table~\ref{table5}. $S_{S}+B_{B}+TL_{M}+D$ can reduce ADE by 13.93$\%$ and 15.85$\%$, and FDE by 17.34$\%$ and 22.46$\%$, respectively. This directly illustrates that the SDG and BDG can effectively capture discontinuous dependency in spatial-temporal space to further improve the accuracy of prediction trajectories.

\textbf{Evaluation of the discriminator:}
We introduce a discriminator to refine the predicted trajectories. By comparing $S_{S}+B_{B}+TL_{M}$ with $S_{S}+B_{B}+TL_{M}+D$ in Table~\ref{table5}, the performances of the latter
are increased by 9.26$\%$ and 12.74$\%$ in ADE and FDE, respectively. Moreover, the discriminator contributes to improving the accuracy of predicted trajectory.

\textbf{Evaluation of different encoders:}
Due to the distinctive characteristics of traffic light states, we use the MLP and LSTM to encode them. By comparing $S_{S}+B_{B}+TL_{L}+D$ and $S_{S}+B_{B}+TL_{M}+D$ in Table~\ref{table5}, utilizing MLP to capture features of traffic light states can be further improved by 5.56$\%$ and 8.17$\%$ on ADE and FDE, respectively. This illustrates that a discontinuous sequence may not be suitable for being encoded by LSTM with strong context correlation.

\textbf{Evaluation of the Function of Traffic Lights:}
For traffic lights, we compare the methods+TL with the corresponding baseline methods. The former directly uses the VTP-TL dataset, and the latter uses a dataset that consists of $Fid$, $Aid$, $x$, and $y$ attributes split from the VTP-TL dataset. As shown in Table~\ref{table3}, it can further increase the performance by 8.02$\%$ to 24.87$\%$ and 3.38$\%$ to 30.29$\%$ in ADE and FDE, respectively. Therefore, we can clearly validate the necessity of traffic lights in trajectory prediction at urban intersections.

\begin{table}[!t]
    \scriptsize
	\begin{center}
	\caption{The ablation results on VTP-TL dataset. \textbf{S} denotes spatial interaction achieved by GAT ($S_{G}$) or SDG ($S_{S}$). \textbf{B} denotes behavior dependency achieved by LSTM ($B_{L}$) or BDG ($B_{B}$). \textbf{TL} denotes traffic light encoder as LSTM ($TL_{L}$) or MLP ($TL_{M}$). \textbf{D} denotes the discriminator. The bold fonts correspond to the best results.}
	
	\label{table5}
	\setlength{\tabcolsep}{1.5mm}{
		\begin{tabular}{l|c|c|c|c|c|c|c|c}
			\hline
			
			\multirow{2}{*}{Setting} & \multicolumn{2}{|c}{\textbf{S}} & \multicolumn{2}{|c}{\textbf{B}} & \multicolumn{2}{|c}{\textbf{TL}} & \multicolumn{1}{|c}{\textbf{D}} & \multicolumn{1}{|c}{Metrics} \\
			
			 \cline{2-9} & GAT & SDG & LSTM & BDG & LSTM & MLP & D& ADE / FDE \\
			 
			 \hline
			 $S_{G}+B_{L}+TL_{M}+D$ & $\surd$ &  & $\surd$ &  &  & $\surd$  & $\surd$ & 21.792 / 48.936 \\
			 \hline
			 $S_{G}+B_{B}+TL_{M}+D$ & $\surd$ &  &  & $\surd$ &  & $\surd$ & $\surd$ & 19.635 / 41.804 \\
			 \hline
			 $S_{S}+B_{L}+TL_{M}+D$ &  & $\surd$ & $\surd$ &  &  & $\surd$ & $\surd$ & 20.082 / 44.560 \\
			 \hline
			 $S_{S}+B_{B}+TL_{L}+D$ &  & $\surd$ &  & $\surd$ & $\surd$ &  & $\surd$ & 17.896 / 37.629 \\
			 \hline
			 $S_{S}+B_{B}+TL_{M}$ &  & $\surd$ &  & $\surd$ &  & $\surd$ &  & 18.626 / 39.598 \\
			 \hline
			 $S_{S}+B_{B}+TL_{M}+D$ &  & $\surd$ &  & $\surd$ &  & $\surd$ & $\surd$ &  \textbf{16.900} / \textbf{34.553} \\
			
			\hline
		\end{tabular}}
	\end{center}
\end{table}

\subsection{Qualitative Evaluation}

\begin{figure}[!t]
  \begin{center}
		\includegraphics[width=0.74\linewidth]{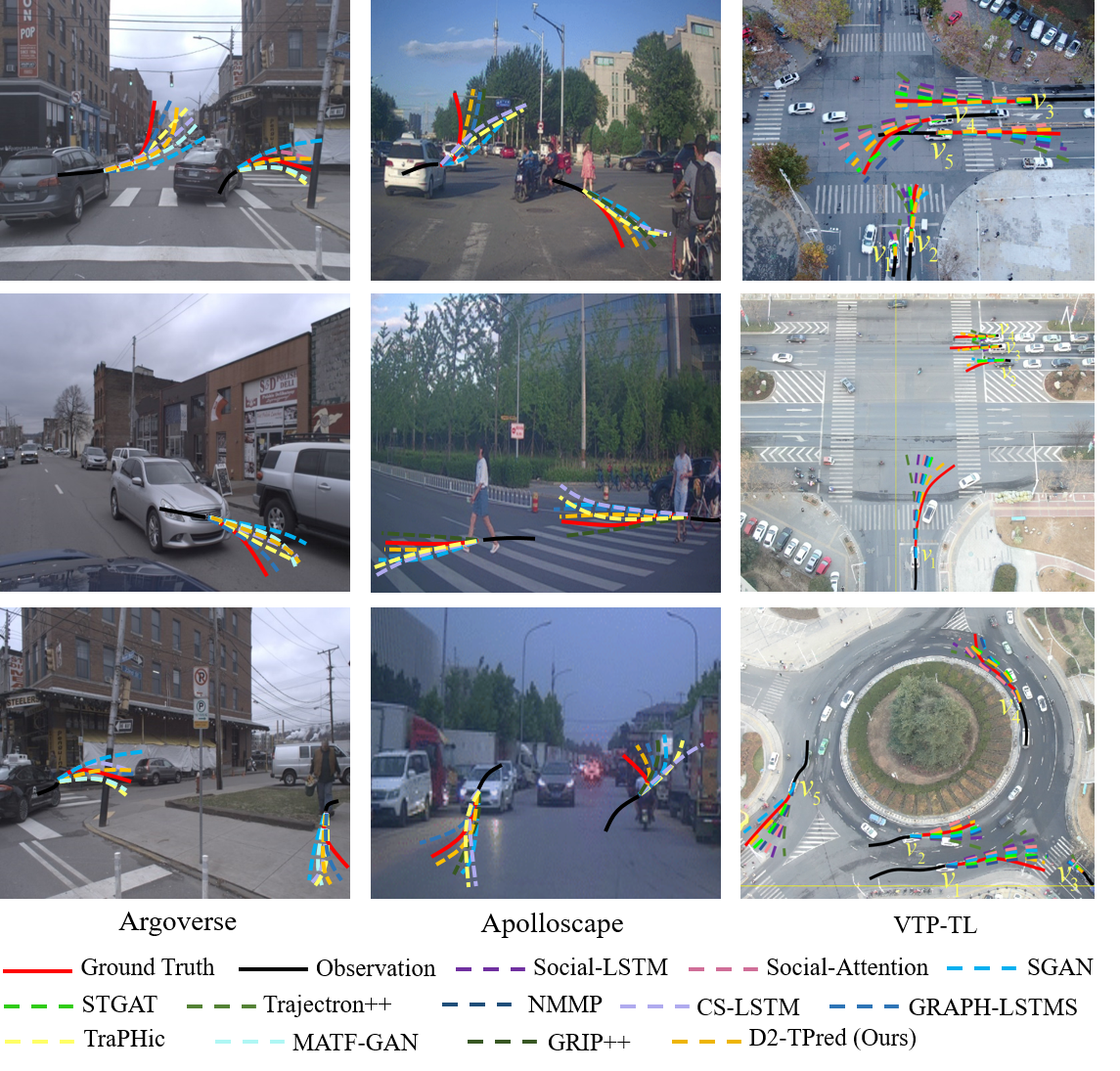}
	\end{center}
  \caption{The visualization results at urban intersections on traffic datasets and VTP-TL dataset. Note that the compared methods are not the same in different datasets.}
  \label{fig5}
\end{figure}

In Figure~\ref{fig5}, the images of first two columns show the qualitative results derived from the Argoverse and Apolloscape. 
It can be seen our method without traffic lights also predicts acceptable future paths at urban intersections. 

In the third column, we show the qualitative results on the VTP-TL dataset. For the first row, the current traffic light state is red on the vertical road. We only show five vehicles' trajectories, where vehicle $v_{1}$ drives straight, $v_{2}$ turns right under the red traffic light, $v_{3}$ goes straight under the green light, $v_{4}$ and $v_{5}$ are not in the influence area of traffic light signals. For $v_{1}$, the predicted trajectory of our method is closest to the ground truth. Although the trajectories of $v_{2}$, $v_{3}$, $v_{4}$, and $v_{5}$ are not affected by traffic light signals, our method also predicts acceptable trajectories. The next two images show the predicted trajectory at T-junctions and roundabout intersections, where the states of vehicles located in the former are changing from parking to driving under the traffic light states from red to green. This illustrates our model can flexibly respond to the dynamic changes of surrounding agents and traffic light states. Limited by the pages, more results and failure case are listed in supplementary materials.

\section{Conclusions}
We present \textit{D2-TPred}, a new trajectory prediction approach by taking into traffic lights.
The approach can not only model the dynamic interactions by reconstructing sub-graphs for all agents with constantly changing interaction objects (SDG), but also captures discontinuous behavior dependency by modeling the direct effects of behaviors at prior instances on the current state (BDG). Moreover, one new dataset VTP-TL for vehicles trajectory prediction with traffic lights is also released. Based on it, we describe two trajectory forecasting schemes and obtain competitive performance against other state-of-the-art.

\textbf{Acknowledgments.} This work was supported in part by the National Natural Science Foundation of China with Grant No.61772474 and 62036010, Zhengzhou Major Science and Technology Project with Grant No.2021KJZX0060-6. We thank all the reviewers for their valuable suggestions.

\clearpage
%
%
\bibliographystyle{splncs04}
\bibliography{egbib}
\end{document}